\documentclass[11pt]{article}

\usepackage{graphicx}
\usepackage{epsfig}
\usepackage{epstopdf}
\usepackage{subfigure}
\usepackage{amsmath}
\usepackage{multirow}
\usepackage{algorithmic}
\usepackage{algorithm}

\begin{document}
\begin{sloppy}

\title{\bf Idealized Dynamic Population Sizing for Uniformly Scaled Problems}
\author{    {\bf Fernando G. Lobo}\\
            \small DEEI-FCT and CENSE\\
            \small Universidade do Algarve\\
            \small Campus de Gambelas\\
            \small 8005-139 Faro, Portugal\\
            \small fernando.lobo@gmail.com
}
\date{}
\maketitle

\begin{abstract}
This paper explores an idealized dynamic population sizing strategy for solving additive decomposable problems of uniform scale. The method is designed on top of the foundations of existing population sizing theory for this class of problems, and is  carefully compared with an optimal fixed population sized genetic algorithm. The resulting strategy should be close to a lower bound in terms of what can be achieved, performance-wise, by self-adjusting population sizing algorithms for this class of problems.
\end{abstract}

\section{Introduction}
\label{sec:intro}

Properly sizing the population of a genetic algorithm (GA) is not an easy task because different problems have different population size requirements. Population sizing theory of GAs tells us that for a given problem, there is a minimum population size below which the GA is not expected to work well~\cite{Goldberg:92,Harik:99b}. The cited works relate important factors that have a direct influence on what should be an adequate population size, and although it resulted in guidelines for setting the population size, they are not immediately applicable to an arbitrary problem because some of those factors (such as signal to noise ratios, building block size, and so on) as well as the conditions under which the model applies (additive decomposable functions of equal or near-equal scale) are rarely met in practice.

Because of that, and also because various researchers believe that no set of static parameter values seems appropriate for an evolutionary algorithm run~\cite{Eiben:07}, various methods have been proposed in the literature to adapt or automate the population size, without requiring the user to specify its value beforehand. Two major categories of methods have been proposed: single and multiple population based approaches. The first category consists of methods that evolve a single population whose  size changes from generation to generation according to some criteria~\cite{Arabas:94,Back:00,Eiben:04,Smith:95,Fernandes:2001,Yu:05b,Cook:2010}. The second category consists of methods that evolve multiple populations of different sizes, but where the size of each population is kept fixed, as in a traditional GA~\cite{Harik:99,Smorodkina:07}. A review of some of the work in this area is available in~\cite{Lobo:07b}.

In some sense, the methods in the second category do not really adapt the population size through the run. What those methods try to do is to find a minimum sufficiently large population that yields a reasonable solution quality for the problem at hand. These approaches can be seen as methods that search for a good fixed-sized population.
The first category of methods is different in the sense that there is only one population whose size changes through the run. We refer to the single population-based approaches as {\em dynamic population sizing} to reflect the notion that the population size changes during the execution of a run. 

%Although there is some evidence that it is beneficial to let the population size change during the run, the benefits of those methods have not be quantified in precise terms. 

Assessing the benefits of dynamic population sizing is difficult for a variety of reasons. First of all, there are several dynamic population sizing methods proposed in the literature, each with its own strengths and limitations. Second, the benefits that can be obtained for a particular class of problems may not be the same as the benefits for another class of problems. Inspired by the work on population sizing theory~\cite{Goldberg:92,Harik:99b}, we try
to make progress by looking at an idealized dynamic population sizing method for additive decomposable problems of uniform scale. The reason for choosing this class of problems is because it is the only class of problems for which theoretical population sizing models exist, and therefore, a fair comparison can be made between a dynamic population sizing method and a traditional GA with a population size set according to the theory.

%As we shall see,  there is indeed some benefits in letting the population size change through the run, but those benefits are not that spectacular when compared with an exact same GA that uses an optimal fixed sized population.

% road map
The rest of the paper is organized as follows. We start in Section~\ref{sec:theory} by reviewing relevant population sizing theory in genetic algorithms, paying close attention to the supply and decision models, and their integration in the gambler's ruin model. Section~\ref{sec:adaptive} builds on the theoretical work to design an idealized strategy for adjusting the population size through a GA run. Section~\ref{sec:experiments} presents experimental results on additive decomposable problems of uniform scale performed with the idealized algorithm, and a comparison of it with a GA using an optimal fixed-sized population. The paper ends with a summary and conclusions.

\section{Population Sizing Theory}
\label{sec:theory}

There are only a few theoretical studies that relate the effect that the population size has in terms of GA performance and solution quality  \cite{Goldberg:92,Harik:99b}. Since GAs are complex to analyze, most of those studies make some simplifying assumptions:  (1) work with selecto-recombinative GAs (no use of mutation), (2) use fixed-length and binary-coded strings, (3) use fixed-size and
non-overlapping populations, and (4) solve additively decomposable functions of equal scale. Although these assumptions are made for computational and analytical tractability, most of them can be relaxed with proper adjustments.

These theoretical models assume that the problem to be solved is additively decomposable in a number of $m$ subfunctions.
Each subfunction maps to $k$ decision variables and corresponds to a partition. Under this definition, it is possible to 
have $2^k$ distinct configurations in a partition, one of which is superior to the others and belongs to the global optimal solution. This superior solution is commonly referred to as a {\em building block} (BB), and the problem can be solved to
optimality by combining the $m$ building blocks, one from each partition, in a single individual.
The models also assume that BBs are neither created nor destroyed by the variation operators; the only source of BBs is in the initial population.  The task of the GA is then to propagate the BBs through selection, and combine them in a single solution using the crossover operator.

Under these assumptions, it has been shown that GAs can efficiently solve this class of problems provided that the variation operators are not too disruptive, and that two crucial aspects are satisfied: (1) there is an adequate {\em supply} of BBs in the initial generation, and (2) the selection operator is able to correctly distinguish between a BB and its competitors. 

\subsection{Supply Models}
\label{sec:supply-models}
When using selecto-recombinative GAs (no mutation), the only source of diversity is the supply of BBs in the initial generation.
A simple supply model considers the number of BBs present in the initial random population (generated under a uniform distribution), where the probability to generate a single BB of size $k$ is $1/2^k$, for binary encodings.
Therefore, the initial supply of BBs on a partition can be estimated as $x_{0}=n/2^k$, where $n$ is the population size.
From this simple relation we observe that the population required to supply BBs grows
exponentially with the BB size $k$. This suggests that problems with short BBs require smaller populations than the 
ones with longer BBs. It also suggests that $k$ has to be much smaller than the whole problem size, otherwise the problem becomes intractable.

\subsection{Decision Models}
\label{sec:decision-models}

The second aspect where population size plays an important role is
in the decision-making between a BB and its competitors. An example is helpful to understand this issue.
Let us consider a competition between two individuals, where for a
given partition of interest, one individual contains a BB and the other individual contains the BB's toughest competitor (i.e, the second best configuration of bits on that partition). The situation is depicted in Figure~\ref{fig:competingBBs} with 
the BB being 1111 and the competitor being 0000.

\begin{figure}
\centering
\includegraphics[width=0.6\linewidth]{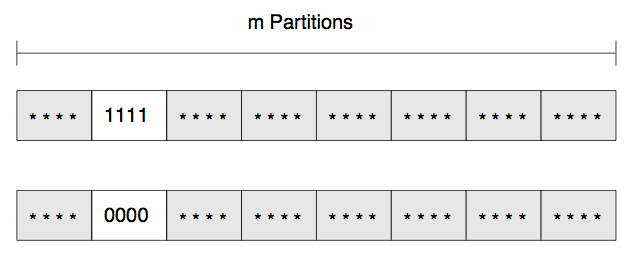}
\caption{In this example, 1111 is a building block and 0000 is its toughest competitor in the second partition.
Although the building block is better than its competitor, there is no guarantee that an individual containing 1111 is better
than an individual containing 0000, because the contents in the remaining partitions act as noise for the correct decision.}
\label{fig:competingBBs}
\end{figure}

During selection, the GA always prefers better individuals. But at the partition level,
it does not always choose the best configuration. This occurs because the remaining
$m-1$ partitions also contribute to the fitness of the individuals in competition, and
act as noise in the decision making process in the particular partition. 

Nonetheless, individuals containing BBs have a selective advantage over those that do not.
By focusing on a single partition alone, Goldberg~et~al.~\cite{Goldberg:92} calculated the probability 
that an individual containing a BB has a greater fitness than an individual that contains the BB's competitor. 
That probability turns out to be

\begin{equation}
p = \Phi \left( \frac{d} {\sqrt{ 2  \sigma_{M}^2} }  \right) ,
\label{eq:pdwell-sigmaM}
\end{equation}

\noindent
where $\Phi$ is the cumulative distribution function for the standard normal distribution,
$d$ if the fitness difference between the BB and its toughest competitor, and
$\sigma_{M}^2$ is the mean fitness variance of the individuals that contain the BB and the individuals that contain the BB's toughest competitor. 

By assuming that the fitness function is the sum of $m$ independent subfunctions, Goldberg~et~al.~note that the 
overall variance of the function can be calculated as the sum of the variances of each independent subfunction. 
By recognizing that the partition of interest does not contribute to the variance, the overall value for $\sigma_{M}^2$
is simply $(m-1) \sigma_{bb}^2$, with $\sigma_{bb}^2$ being the subfunction variance. The probability of deciding well between a BB and its closer competitor, on a single trial, can then be written as

\begin{equation}
p = \Phi \left( \frac{d} { \sqrt{2 (m-1)} ~ \sigma_{bb}} \right) .
\label{eq:pdwell}
\end{equation}

Based on this result, Goldberg~et~al.~\cite{Goldberg:92} were able to derive a population sizing equation which gave
conservative estimates for the convergence quality of the GA on additive decomposable problems with uniformly scaled subfunctions. The resulting equation gave the required population size
(in order to correctly solve each partition with a tolerated error probability $\alpha$) in terms of the
BB size $k$, the total number of partitions $m$, and the fitness difference $d$ between a BB and its toughest competitor.
A few years later, Harik~et~al.~\cite{Harik:99b} proposed another model that incorporates the initial BB supply as well as cumulative effects of decision making over an entire GA run. Their model is based on the
well-known gambler's ruin problem and is discussed in more detail in the next subsection.

\subsection{The Gambler's Ruin Model}
\label{sec:gr-model}

Harik~et~al.~\cite{Harik:99b} modeled the behavior of the GA as one dimensional random walks. Their work puts together the previous models of building block supply and decision making into a single integrated model, and resulted in a more accurate prediction for the convergence quality of the GA. 

Given the assumption that the partitions are independent, the authors were able to focus on the convergence of a single partition alone. In order to do so, they made an analogy between a bounded one-dimensional random walk and the action of selection in GAs. 

Consider a population of size $n$. For a given partition, the GA succeeds when it has $n$ copies of the BB in that partition
(meaning the GA has fully converged to the correct BB in that partition), and fails when it has 0 copies of the BB (meaning that the GA lost the BB and is not able to recover it). 
In the initial generation, the number of BBs is given by the supply model, $n/2^k$. 
Then, throughout the GA run, the number of BBs can increase or decrease depending on whether the selection operator is able to make correct decisions between a BB and its competitors. Eventually, the GA either succeeds or fails.

The number of BBs can be represented by the position, $x$, of a particle in a one-dimensional random walk with two absorbing barriers at $x=0$ (BB lost) and $x=n$ (BB fully propagated). The absorbing barriers are due to the fact that the model ignores the effects of BB creation and destruction by the variation operators. Since the decision errors done by selection are inevitable, the population size $n$ should be large enough to be able to tolerate some of these errors. 

\begin{figure}
\centering
\includegraphics[width=0.6\linewidth]{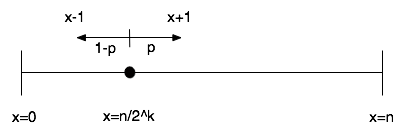}
\caption{The gambler's ruin problem. At each time step, the particle $x$ either moves one unit to the right, with probability $p$, or one unit to the left, with probability $1-p$.}
\label{fig:random-walk}
\end{figure}

In order to predict the BB success or failure, Harik~et~al.~abandoned the notion of generations in a GA and considered that selection makes a sequence of decisions, one at a time, between a BB and a competitor, until the BB is either lost or fully propagated in the partition of interest. Under this view, the outcome of a single competition is to win or loose a single copy of the BB, and the probability of success can be calculated using the solution to the gambler's ruin problem~ \cite{Feller:66}. This problem is a special case of a bounded one-dimensional random walk, where at each time step, the particle $x$ either moves one unit to the right (with probability $p$) or moves one unit to the left (with probability $q=1-p$). Figure~\ref{fig:random-walk} illustrates the problem. The solution to gambler's ruin problem is given by Equation~\ref{eq:psuccess}. It gives the probability of success as a function of the starting position of the particle, $x=x_0$, and the probability $p$ of making a step in the right direction.
\begin{equation}
P_{success} = \frac{1-\Big(\frac{q}{p}\Big)^{x_{0}}}{1-\Big(\frac{q}{p}\Big)^{n}}
\label{eq:psuccess}
\end{equation}

It is illustrative to observe the effect of $n$ in the gambler's ruin model. As an example, let $p=0.51$,
$n=10$, and $x_0 = n/2 = 5$. Plugging these values in Equation~\ref{eq:psuccess} gives $P_{success} \approx 0.5498$.
If we instead let $n=1000$ and $x_0 = n/2 = 500$, the probability of success becomes $P_{success} \approx 0.9999$.
By raising $n$ (and $x_0$ in the same proportion), the probability of success increases, which is in agreement with the notion that larger populations have a better chance of obtaining high quality solutions in a selecto-recombinative GA.

The analogy between the gambler's ruin problem and the convergence behavior of a GA should be clear now.
The probability $p$ of making a decision in the right direction is simply the probability of deciding well, given by Equation~\ref{eq:pdwell}, and the starting position $x_0 = n/2^k$ corresponds to the initial supply of BBs in a randomly initialized population.

By equating the probability of success to $1-\alpha$ (where $\alpha$ is the failure rate, the allowed probability that the GA looses the BB) in Equation~\ref{eq:psuccess}, and solving for $n$, Harik~et~al.~were able to find the population size required to correctly solve a partition, given the specified failure rate $\alpha$. The resulting equation is
\begin{equation} 
  n ~ =  -2^{k-1}  \ln (\alpha)  ~ \frac{\sigma_{bb}  \sqrt{\pi (m-1)}}{d},
\label{eq:harik}
\end{equation}

\noindent
where the variables $k$, $d$, $m$, and $\sigma_{bb}$ are as discusses in the previous subsections.
Under the assumption that a problem consists of $m$ independent subfunctions of uniform scale, 
a GA with a population sized according to Equation~\ref{eq:harik} is expected to correctly solve an average of $m (1-\alpha)$ subfunctions.
Experimental results from the original paper show that the equation gives a reliable estimate for population sizing on decomposable
problems with uniformly-scaled and near-uniformly-scaled building blocks. 

\section{Designing an Idealized Dynamic Population Sizing Algorithm}
\label{sec:adaptive}

The gambler's ruin model gives an accurate estimate for the minimum population size required to solve additive decomposable problems of uniform scale with a desired minimum accuracy (resp. maximum error). The resulting population size can then be used by the GA. We note however that the resulting GA has a static population size, albeit a properly sized one, and as some researchers have suggested, a static parameter value, no matter what the value is, is unlikely to be the very best choice. We note that this argument is debatable as any reasonable adaptive algorithm is also unlikely to be optimal since in order to properly adapt, the algorithm has to make an effort to learn a correct adaptation strategy, and it is well known that any learning algorithm has a cost.

Philosophical arguments aside, we proceed with a fundamental question: Is there a dynamic population sizing strategy for additive decomposable problems of uniform scale, that is able to reach the same solution quality as a GA with an optimal fixed population size, but do so using fewer fitness function evaluations? If yes, what are the speedups that can be achieved with such a strategy? These are the question that we explore in the rest of this paper. The reason for choosing this class of problems is because the existing population sizing theory is immediately applicable, and as we shall see, the theory is helpful in designing an idealized dynamic population strategy. A word of caution should be said though. The resulting strategy assumes knowledge about the problem, just like the existing population sizing theory also assumes knowledge of the problem, and therefore, it is not applicable to an arbitrary problem. Nonetheless, studying  idealized dynamic population sizing for this class of problems is an important topic as it can give insights into the limits that can be achieved in terms of performance. 

An ideal dynamic population size algorithm for this class of problems should pay close attention to the theory of population sizing. The algorithm should choose, at any given point in time, its best guess for the adequate population size. One way of doing this is by paying a close attention to the factors than influence BB propagation. Our previous discussion on supply and decision models, and their combined integration in the gambler's ruin model, shed some light into the variables that can be controlled in order to achieve an idealized strategy. The variables of interest are the building block supply and the probability of deciding well between a BB and its closer competitor. These variables are calculated once and remain static in the gambler's ruin model. But on a real GA run, they change. 

\subsection{Better decisions with less variance}
\label{sec:varfit}

The gambler's ruin model assumes a constant probability $p$ of deciding correctly between a BB and its tougher competitor. In a real GA run, this probability is not constant for two reasons. First, the BB does not always compete with its closer competitor, even though after a few generations,  most of the competitions will involve the two of them, and therefore this reason can be safely ignored. Second, and more important, the fitness variance of the population changes through time. It is typically large at the beginning of the run, and decreases as the GA converges. This means that the probability of deciding well between a BB and its competitor should be higher later in the run, and that translates into not needing such a large population size as in the beginning of the search. An example should clarify this issue. Consider a problem with partitions of size $k=1$, and a competition between a BB and its competitor at a particular partition of interest,
\begin{center}
1*****\\
0*****\\
\end{center}

The correct decision between a member of 1***** and a member of 0***** is, as discussed in section~\ref{sec:decision-models}, blurred by the contents of the non-fixed positions. Earlier in the run, the contents of those non-fixed positions are more or less uniformly distributed, which means that when we compare two strings bit by bit, roughly half of the positions should contain a different bit value.  As the search progresses, the contents of these non-fixed positions start themselves to converge and become further and further away from being uniformly distributed. As an extreme case, when all the non-fixed positions fully converge, the correct decision between the BB and its competitor (in the first partition of our example) can be made with absolute certainty.

In the gambler's ruin model, as well as in the earlier population sizing model~\cite{Goldberg:92}, the probability of deciding well is considered to be fixed during the entire run, and is calculated assuming the contents of the non-fixed positions are uniformly randomly distributed (i.e., the overall variance coming from the $m-1$ partitions is considered to be $(m-1) \sigma_{bb}^2$).
This is a reasonable assumption to make in the model because it needs to give a conservative estimate for the population size. 
The fact that the variance does change in a real GA run should however be explored when designing dynamic population sizing strategies, and indeed it has been done in the past \cite{Smith:95,Yu:05b}.

A simple way to have a dynamic population sizing strategy is to adjust the probability $p$ of deciding well between a BB and its competitor in the gambler's ruin model. To do so, we can calculate $p$ from Equation~\ref{eq:pdwell-sigmaM} rather than from Equation~\ref{eq:pdwell}, and use for $\sigma_{M}^2$ the population fitness variance. This way, the gambler's ruin model can be used anew in each generation with a different probability $p$. 
Once $p$ is calculated, the population size for the next generation can be estimated as
\begin{equation} 
  n ~ = ~ \frac{2^k \ln (\alpha)}{ \ln \left( \frac{1-p}{p} \right)},
\label{eq:harik-p}
\end{equation}

Equation~\ref{eq:harik-p}  was also obtained by Harik~et~al.~\cite{Harik:99b} as an intermediate step to derive Equation~\ref{eq:harik}. 
This thought experiment suggests that such a dynamic strategy should be able to  solve the same class of additive decomposable problems given the same tolerated $\alpha$ failure rate. After all, it is still obeying the conditions of the model. 
The result should be a less conservative estimate for BB success because the population is adjusted at each generation to the minimum size required in order to make good decisions between a BB and its competitors. It remains to see if it does so with less fitness function evaluations than a GA that has its populations size statically set according to Equation~\ref{eq:harik}.

\subsection{Changes in building block supply}
\label{sec:varfit-x0}

The strategy that was outlined for adapting the population size can still be made more tight. Our discussion so far suggests that we use a different value for $p$ in Equation~\ref{eq:harik-p} by monitoring the fitness variance of the population through time. The BB supply however is still considered to be static and fixed to $n/2^k$. But in an actual run, the BB supply also changes from generation to generation. If we consider an idealized situation where the algorithm can monitor the number of BBs in each partition at any given generation, then we could use that information to use a different value for $x_0$ as the search progresses. The problem is that as the search progresses, the partitions cannot be considered independent with respect to the BB supply (i.e., it does not make sense to take the average number of BBs per partition as an overall value for the supply). In reality, we have $m$ random walks taking place, one for each partition, and in each partition the particle denoting the current number of BBs can be at a different position. The strategy that we follow, is to choose for the building block supply the maximum between $n/2^k$ and the number of BBs present in the $\alpha$-percentile worst partition in terms of number of BBs. By using the $\alpha$-percentile we are in some sense telling the GA that we want to have a population size large enough to correctly solve that partition, but do not mind failing to solve partitions whose supply of BBs is worse than that.
Also, the reason for not letting the supply go below $n/2^k$ is because if the BB supply gets too low, the resulting population size may become prohibitively large, and we might as well loose the BB. As we shall see, this heuristic seems to be a reasonable choice, and still gives a conservative estimate for the solution quality of the GA. The next section presents some experiments to validate these ideas.

\section{Experiments}
\label{sec:experiments}

\subsection{Test functions}

We performed a set of experiments on two additive decomposable problems of uniform scale: the onemax function and the concatenated trap function with 4 bits per trap. These two functions were also used in~\cite{Harik:99b}. 
In onemax the fitness is given by the number of ones in a binary string $X$:
\begin{equation}\label{eq:onemax}
 f_{onemax}(X) = \sum_{i=0}^{\ell-1} x_i.
\end{equation}

For the concatenated trap function, the fitness is given by the sum of the fitness contributions of $m$ subfunctions, each a trap function defined over 4 variables.
\begin{equation} 
  f_{m-trap4}(X) = \sum_{i=0}^{m-1} f_{trap4}(x_{4i},x_{4i+1},x_{4i+2},x_{4i+3}).
\end{equation}

The fitness contribution of each trap function is obtained by first calculating the number of ones, $u$, contained in the substring of 4 variables. Then the fitness is given by
\begin{equation}\label{eq:12}
  f_{trap4}(u) = \left\{ \begin{array}{ll}
      4,      & \textrm{if $u$ = $4$}\\
      3-u,  & \textrm{otherwise}
    \end{array} \right.
\end{equation}

With onemax we conduct experiments with string lengths 100, 200, 300, 400.
With the concatenated trap functions, we do experiments with $m = 20, 40, 60, 80$ subfunctions.
In onemax the partitions are of size $k=1$, and in the concatenated trap function the partitions are of size $k=4$.

\subsection{GA setup}

We use a generational GA with full replacement, binary tournament selection without replacement, and no mutation, precisely as described in the experiments done in~\cite{Harik:99b}. For the crossover operator, however, we simply shuffle the population partition by partition. This is the equivalent of sampling a new population based on the observed frequencies of the various partition configurations, as done in Estimation of Distribution Algorithms with known factorizations \cite{Muhlenbein:99}. 
We opt to use this kind of operator in order to stay closer to the assumptions of the gambler's ruin model (i.e, that the partitions are completely decorrelated). This sampling procedure is the equivalent of performing a kind of uniform crossover at the partition level, either exchanging or not exchanging a chunk of $k$ linked bits, but doing so an infinite number of times per generation. 

We also take care in not evaluating the same solution more than once in an entire run. We achieve this by using an archive of visited solutions. We first check if it is in the archive. If yes, we simply retrieve its value. Otherwise, we evaluate it and keep it in the archive along with its fitness value for future use. This is a sensible strategy to do in real GA application because fitness evaluation can be very time consuming, although that is not the case in our artificial test functions. The reason for using the archive in our experiments is for not giving any sort of disadvantage to the static population sized GA, since later in the run, when most of the population has converged more or less to the same solutions, a GA with a larger population is more likely to be re-evaluating the same solution over and over. 

The only difference in the GAs is in their population sizing strategy. Three GAs are used. One has a static population size set according to Harik~et~al.'s gambler's ruin model (Equation~\ref{eq:harik}), the other two use a dynamic population sizing strategy as explained in sections~\ref{sec:varfit} and~\ref{sec:varfit-x0}, respectively. For the initial population, the GAs with dynamic population size also start with the population size dictated by Equation~\ref{eq:harik}. After estimating the population size for the next generation, the new population is created from the current one by applying the GA operators until the desired population size is reached. The estimated population size is always rounded up to the next even integer. A minimum population size of 4 individuals is enforced at all times. 
A failure rate $\alpha=0.05$ is used in all experiments, meaning that all GAs are expected to correctly solve 95\% of the partitions (or subfunctions). 

For each problem instance, and for each algorithm, we perform 100 independent runs. 
Each run ends when all BB partitions converge, i.e., for every partition, either the BB is fully propagated or lost entirely.
At the end of each run, we record the number of fitness function evaluations and the solution quality in terms of proportion of subfunctions correctly solved. 

\subsection{Results}
\label{sec:results}

\subsubsection{Onemax experiments}

Figure~\ref{fig:onemax1} presents the results obtained on the onemax function for different problem sizes.
On the left, the number of fitness function evaluations taken by each algorithm is plotted. On the right,
the solution quality obtained is plotted. 
(On all plots, the key ``dynamic varfit" refers to the GA that adapts the probability $p$ of deciding well according to the fitness variance of the population as discussed in section~\ref{sec:varfit}, and the key ``dynamic varfit \& supply" refers to the GA that adapts both the probability $p$ as well as the building block supply $x_0$, as discussed in section~\ref{sec:varfit-x0}.)

\begin{figure*}[t!]
  \center
  \subfigure[fitness evaluations]{\includegraphics[width=0.45\linewidth]{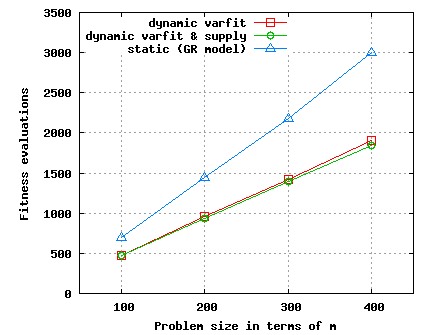}}
  \subfigure[solution quality]{\includegraphics[width=0.45\linewidth]{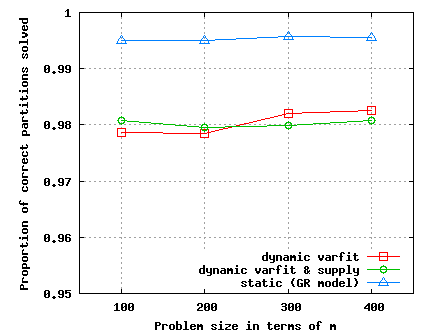}}
  \caption{Fitness function evaluations and solution quality obtained by the various algorithms on the onemax problem. }
  \label{fig:onemax1}
\end{figure*}

\begin{figure*}[t!]
  \center
  \subfigure[speedup ratio]{\includegraphics[width=0.45\linewidth]{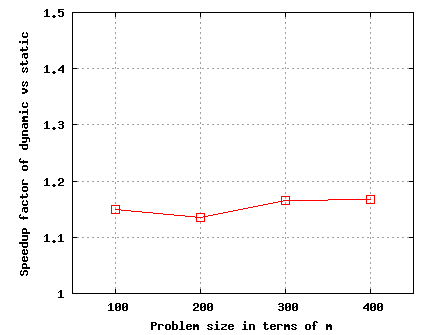}}
  \subfigure[population size through time]{\includegraphics[width=0.45\linewidth]{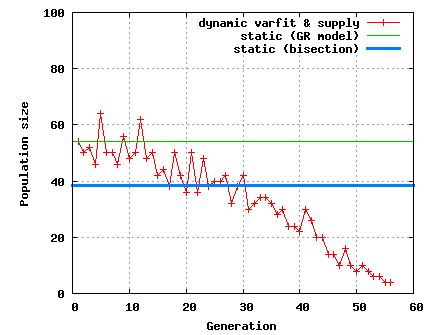}}
  \caption{The left plot shows the speedup ratio of the dynamic population sizing strategy over the optimal fixed-sized population given by the bisection method to reach the same solution quality, for various sizes of the onemax problem. The right plot illustrates the evolution of the population size on a single run of a 400-bit onemax problem. }
  \label{fig:onemax2}
\end{figure*}

The static population sized GA is the slowest of the three but it is able to reach a slightly better solution quality. There is no significant difference between the two dynamic approaches, both in terms of fitness evaluations and accuracy of the solution.
Another important observation is that both dynamic algorithms are able to reach the desired minimum accuracy of $1-\alpha = 0.95$. (Notice that the y-axis in Figure~\ref{fig:onemax1}-(b) ranges from 0.95 to 1.) This is expected because the dynamic strategies were designed taking into account the gambler's ruin model with a specified $\alpha$ probability. 

Based on these results, one could say that the dynamic strategies are better than the fixed-sized GA set according to the population sizing theory. Such a comparison however is not fair because the static sized GA is able to reach a better solution quality than the others. This occurs because the gambler's ruin model still gives a conservative estimate for the minimum population size required to solve the problem. 

In order to make a fair comparison we need to find the minimum fixed-sized population that is able to reach the same solution quality reached by the dynamic population sized strategies. We do so for the case of the algorithm that adapts according to both the variance and supply. In order to find the minimum population size required to reach the desired solution quality, the bisection method~\cite{Sastry:01a,Pelikan:05} over the population size is performed. The results are averaged over 20 independent bisection runs. In each bisection run, the minimum population size needed to reach the target solution, as well as the number of fitness function evaluations taken to do so, are averaged over 50 independent runs. 

The results for the number of fitness function evaluations needed by the GA with a static population given by the bisection method were collected and plotted in Figure~\ref{fig:onemax2}-(a) in terms of speedup ratios. The comparison between the idealized dynamic population sizing strategy and the ``optimal" static population sized GA can now be made on more fair grounds since we are comparing the number of fitness evaluations needed by both algorithms to reach a similar solution quality. It turns out that the dynamic strategy is better than the fixed strategy with an approximate speedup of 15\%.

The plot in Figure~\ref{fig:onemax2}-(b) shows the evolution of the population size on a sample 400-bit onemax run (one of the 100 independent runs performed), along with the fixed population size given by the gambler's ruin model, and the population size given by the bisection method that is able to reliably reach the same solution quality as the dynamic strategy.

\subsubsection{Concatenated trap experiments}

Figures~\ref{fig:trap1} and~\ref{fig:trap2} are the equivalent of 
Figures~\ref{fig:onemax1} and~\ref{fig:onemax2}, for the experiments performed with the concatenated trap functions. 

\begin{figure*}[t!]
  \center
  \subfigure[fitness evaluations]{\includegraphics[width=0.45\linewidth]{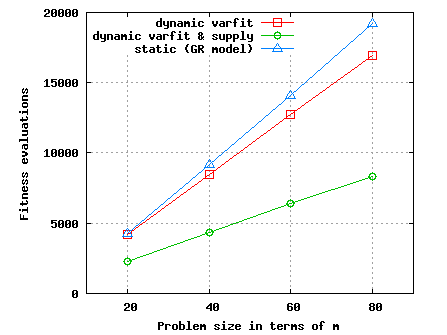}}
  \subfigure[solution quality]{\includegraphics[width=0.45\linewidth]{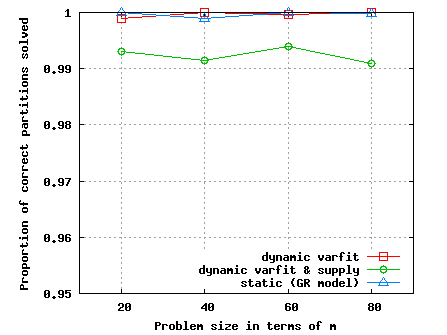}}
  \caption{Fitness function evaluations and solution quality obtained by the various algorithms on the concatenated trap4 function. }
  \label{fig:trap1}
\end{figure*}

\begin{figure*}[t!]
  \center
  \subfigure[speedup ratio]{\includegraphics[width=0.45\linewidth]{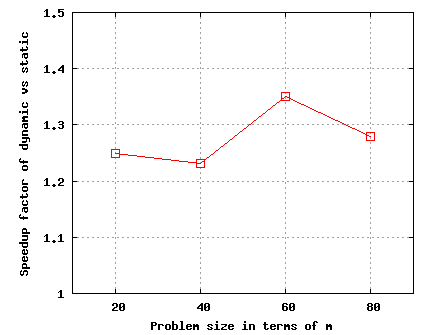}}
  \subfigure[population size through time]{\includegraphics[width=0.45\linewidth]{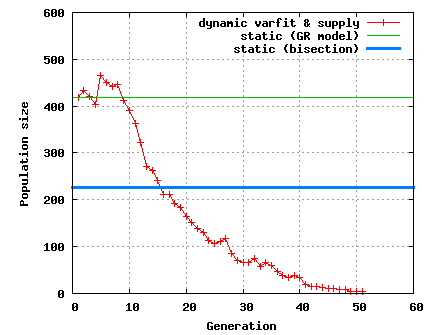}}
  \caption{The left plot shows the speedup ratio of the dynamic population sizing strategy over the optimal fixed-sized population given by the bisection method to reach the same solution quality, for various sizes of the concatenated trap4 function. The right plot illustrates the evolution of the population size on a single run of a concatenated trap4 function with 80 subfunctions (320 bits).}
  \label{fig:trap2}
\end{figure*}

It can be observed that the behavior of the two dynamic population sizing algorithms differ a little bit from each other, as opposed to what happened with the onemax function. In this case, the method that adapts according to the fitness variance and to the BB supply, seems to be better. Notice again that all the algorithms solve the problem with the desired minimum accuracy of 95\%. Indeed, they are all able to solve with an accuracy above 99\%. Again, we performed the bisection method in order to make a fair comparison between the dynamic strategy (varfit \& supply) and a GA with an optimal fixed-sized population. The speedup factors are now in the range of 23-35\%, a little larger than the ones obtained with the onemax experiments.

The observed difference between the two adaptive strategies, as well as the increase in speedup relative to the onemax experiments, can be explained by the fact that dynamically adjusting the supply, $x_0$, in the case of the trap experiments is more beneficial. Remember that in the non-dynamic case, the supply $x_0 = n/2^k$ is kept fixed. For the onemax experiments, $k=1$, and so the supply is half way between the absorbing barriers of the random walk model. In the trap4 experiments however, the supply ($n/16$) is much closer to the failure barrier. Once the GA starts going in the right direction, the initial supply of $n/16$ becomes a very conservative estimate, and an algorithm that is able to adjust its value is likely to obtain more benefits from it, as we indeed observe.

\section{Summary and Conclusions}

This paper addressed the topic of dynamic population sizing in genetic algorithms. By paying close attention to the existing theory of population sizing for the class of additive decomposable problems of uniform scale, we were able to design an idealized dynamic population sizing strategy for this class of problems.

Several researchers have alluded to the benefits of parameter control in evolutionary algorithms and to the limitations of EAs that use a static set of parameter values. Although there is some evidence regarding the benefits of parameter control, little concrete evidence has been reported in terms of the population size parameter for classes of problems where good fixed populations sizes are known to give reliable results in terms of solution quality. Moreover, most of the existing literature on  self-adjusting population sizing, report results from algorithms that are not all able to reach the same solution quality on the given test function set, which makes the comparison difficult to assess. This paper approaches the problem from a different angle. An idealized dynamic population sizing strategy is designed for reaching a given solution quality. Then we search for an optimally fixed population size GA that is able to reach the same solution quality as the one reached by the idealized dynamic strategy, making the comparison more straightforward.

The results presented in this paper do give evidence that there can be benefits in dynamically adjusting the population size, at least for the class of problems tested. More important, since the dynamic method was designed based on well grounded population sizing fundamentals, derived from the theory of GAs for additive decomposable problems, we should expect the resulting strategy to be relatively close to a lower bound in terms of what can be achieved, performance-wise, by a self-adjusting population sizing method for this class of problems.

Finally, the idealized strategy presented in the paper is not intended to be a dynamic population sizing algorithm to be used in practice. It was designed and presented only as an idealized strategy in order to give insights in terms of what can be achieved by population (re)sizing methods. The strategy is of course not applicable in practice because it relies on several assumptions that do not usually hold for real-world problems, as well as on parameters that are either unknown or hard to estimate.

\section*{Acknowledgments}
This work was sponsored by the Portuguese Foundation for Science and
Technology under grant PTDC-EIA-67776-2006.

\bibliographystyle{abbrv}
\bibliography{references}

\begin{thebibliography}{10}

\bibitem{Arabas:94}
J.~Arabas, Z.~Michalewicz, and J.~Mulawka.
\newblock {GAVaPS} -- a genetic algorithm with varying population size.
\newblock In {\em Proc. of the {F}irst {IEEE} {C}onf. on {E}volutionary
  {C}omputation}, pages 73--78, Piscataway, NJ, 1994. IEEE Press.

\bibitem{Back:00}
T.~B{\"{a}}ck, A.~E. Eiben, and N.~A.~L. van~der Vaart.
\newblock An empirical study on {GA}s without parameters.
\newblock In {\em Parallel Problem Solving from Nature, PPSN VI, LNCS 1917},
  pages 315--324. Springer, 2000.

\bibitem{Cook:2010}
J.~E. Cook and D.~R. Tauritz.
\newblock An exploration into dynamic population sizing.
\newblock In {\em Proceedings of the {G}enetic and {E}volutionary {C}omputation
  {C}onference {GECCO}-2010}, pages 807--814. ACM, 2010.

\bibitem{Eiben:04}
A.~E. Eiben, E.~Marchiori, and V.~A. Valko.
\newblock Evolutionary algorithms with on-the-fly population size adjustment.
\newblock In X.~Yao et~al., editors, {\em Parallel Problem Solving from Nature
  {PPSN VIII}, {LNCS} 3242}, pages 41--50. Springer, 2004.

\bibitem{Eiben:07}
A.~E. Eiben, Z.~Michalewicz, M.~Schoenauer, and J.~E. Smith.
\newblock Parameter control in evolutionary algorithms.
\newblock In F.~G. Lobo, C.~F. Lima, and Z.~Michalewicz, editors, {\em
  Parameter Setting in Evolutionary Algorithms}, pages 19--46. Springer, 2007.

\bibitem{Feller:66}
W.~Feller.
\newblock {\em An introduction to probability theory and its applications},
  volume~1.
\newblock John Wiley and Sons, New York, NY, 2nd edition, 1966.

\bibitem{Fernandes:2001}
C.~Fernandes and A.~Rosa.
\newblock A study on non-random mating and varying population size in genetic
  algorithms using a royal road function.
\newblock In {\em Proceedings of the 2001 Congress on Evolutionary Computation
  CEC2001}, pages 60--66. IEEE Press, 2001.

\bibitem{Goldberg:92}
D.~E. Goldberg, K.~Deb, and J.~H. Clark.
\newblock Genetic algorithms, noise, and the sizing of populations.
\newblock {\em Complex {S}ystems}, 6:333--362, 1992.

\bibitem{Harik:99b}
G.~Harik, E.~Cant{\'{u}}-Paz, D.~E. Goldberg, and B.~L. Miller.
\newblock The gambler's ruin problem, genetic algorithms, and the sizing of
  populations.
\newblock {\em Evolutionary Computation}, 7(3):231--253, 1999.

\bibitem{Harik:99}
G.~R. Harik and F.~G. Lobo.
\newblock A parameter-less genetic algorithm.
\newblock In W.~Banzhaf et~al., editors, {\em Proceedings of the {G}enetic and
  {E}volutionary {C}omputation {C}onference {GECCO}-99}, pages 258--265, San
  Francisco, CA, 1999. Morgan Kaufmann.

\bibitem{Lobo:07b}
F.~G. Lobo and C.~F. Lima.
\newblock Adaptive population sizing schemes in genetic algorithms.
\newblock In F.~G. Lobo, C.~F. Lima, and Z.~Michalewicz, editors, {\em
  Parameter Setting in Evolutionary Algorithms}, pages 185--204. Springer,
  2007.

\bibitem{Muhlenbein:99}
H.~M{\"{u}}hlenbein and T.~Mahning.
\newblock {FDA} - {A} scalable evolutionary algorithm for the optimization of
  additively decomposed functions.
\newblock {\em Evolutionary Computation}, 7(4):353--376, 1999.

\bibitem{Pelikan:05}
M.~Pelikan.
\newblock {\em {H}ierarchical {B}ayesian {O}ptimization {A}lgorithm: {T}oward a
  New Generation of Evolutionary Algorithms}.
\newblock Springer, 2005.

\bibitem{Sastry:01a}
K.~Sastry.
\newblock Evaluation-relaxation schemes for genetic and evolutionary
  algorithms.
\newblock Master's thesis, University of Illinois at Urbana-Champaign, Urbana,
  IL, 2001.

\bibitem{Smith:95}
R.~E. Smith and E.~Smuda.
\newblock Adaptively resizing populations: Algorithm, analysis, and first
  results.
\newblock {\em Complex Systems}, 9:47--72, 1995.

\bibitem{Smorodkina:07}
E.~Smorodkina and D.~R. Tauritz.
\newblock Greedy population sizing for evolutionary algorithms.
\newblock In {\em Proceedings of the IEEE Congress on Evolutionary Computation,
  CEC 2007}, pages 2181--2187. IEEE, 2007.

\bibitem{Yu:05b}
T.-L. Yu, K.~Sastry, and D.~E. Goldberg.
\newblock Online population size adjusting using noise and substructural
  measurements.
\newblock In {\em Proceedings of the IEEE International Conference on
  Evolutionary Computation}, pages 2491--2498, 2005.

\end{thebibliography}

\end{sloppy}
\end{document}